\documentclass{article}
\usepackage{amsmath}
\usepackage{amsfonts,amssymb}
\usepackage{amsthm}
\DeclareSymbolFontAlphabet{\mathbb}{AMSb}
\usepackage[numbers]{natbib}
\usepackage{float}
\usepackage{caption}       

\usepackage{fancyhdr}
\usepackage{fancybox}
\usepackage{url}
\usepackage{xspace}
\usepackage{array}
\usepackage{multirow}
\usepackage[mathscr]{euscript}
\usepackage{subcaption}
\usepackage{bookmark}
   
\usepackage{xcolor}
\usepackage{color}
\usepackage{pgfplots}
\usepackage{pgfplotstable}
\usepackage{hhline}

\usepackage{graphicx}
\usepackage{epstopdf}

   \usepackage[preprint]{neurips_2020}

\usepackage[utf8]{inputenc} 
\usepackage[T1]{fontenc}    
\usepackage{hyperref}       
\usepackage{url}            
\usepackage{booktabs}       
\usepackage{amsfonts}       
\usepackage{nicefrac}       
\usepackage{microtype}      

\newcommand{\vertiii}[1]{{\left\vert\kern-0.25ex\left\vert\kern-0.25ex\left\vert #1 
		\right\vert\kern-0.25ex\right\vert\kern-0.25ex\right\vert}}
\newcommand{\be}[1]{\begin{equation}\label{#1}}
\newcommand{\benon}{\begin{equation*}}  
\newcommand{\bemuln}[1]{\begin{multline}\label{#1}}
\newcommand{\bemul}{\begin{multline*}}
\newcommand{\bee}{\begin{eqnarray*}}
\newcommand{\eee}{\end{eqnarray*}}
\newcommand{\been}[1]{\begin{eqnarray}\label{#1}}
\newcommand{\eeen}{\end{eqnarray}}
\newcommand{\began}[1]{\begin{gather}\label{#1}}
\newcommand{\bega}{\begin{gather*}}
\newcommand{\bealn}[1]{\begin{align}\label{#1}}
\newcommand{\beal}{\begin{align*}}
\newcommand{\bealatn}[2]{\begin{alignat}{#1}\label{#2}}
\newcommand{\bealat}{\begin{alignat*}}
\newcommand{\bexalatn}[1]{\begin{xalignat}\label{#1}}
\newcommand{\bexalat}{\begin{xalignat*}}

\newcommand{\mbb}{\mathbb}

{\theoremstyle{plain} 
\newtheorem{thm}{Theorem}[section]

}
{\theoremstyle{break} 
\newtheorem{defi}{Definition}
\newtheorem{ass}{Assumption} 
 
}

\def\ba{{\mathbf a}}
\def\bb{{\mathbf b}}

\def\bv{{\mathbf v}}
\def\bw{{\mathbf w}}
\def\bx{{\mathbf x}}  
\def\by{{\mathbf y}}
\def\bz{{\mathbf z}}
\def\bA{{\mathbf A}}
\def\bB{{\mathbf B}}

\def\bI{{\mathbf I}}

\def\bY{{\mathbf Y}}

\def\bW{{\mathbf W}}

\def\texitem#1{\par\smallskip\noindent\hangindent 25pt
               \hbox to 25pt {\hss #1 ~}\ignorespaces}

\newcommand{\scrC}{\mathcal{C}}
\newcommand{\scrD}{\mathcal{D}}

\newcommand{\scrN}{\mathcal{N}}

\newcommand{\scrP}{\mathcal{P}}

\newcommand{\scrZ}{\mathcal{Z}}

\newcommand{\bbeta}{\boldsymbol{\beta}}

\newcommand{\btheta}{\boldsymbol{\theta}}

\newcommand{\bSigma}{{\boldsymbol{\Sigma}}}

\title{Robustified Multivariate Regression and Classification Using Distributionally Robust Optimization under the Wasserstein Metric}

\author{%
	Ruidi Chen\\
	Division of Systems Engineering\\
	Boston University\\
	Boston, MA 02215 \\
	\texttt{rchen15@bu.edu} \\
	\And
	Ioannis Ch. Paschalidis \thanks{http://sites.bu.edu/paschalidis}  \\
	Department of Electrical and Computer Engineering \\
	Division of Systems Engineering \\
	and Department of Biomedical Engineering \\
	Boston University\\
	Boston, MA 02215 \\
	\texttt{yannisp@bu.edu} \\
}

\begin{document}

\maketitle

\begin{abstract}
  We develop {\em Distributionally Robust Optimization (DRO)} formulations for {\em Multivariate Linear Regression (MLR)} and {\em Multiclass Logistic Regression (MLG)} when both the covariates and responses/labels may be contaminated by outliers. The DRO framework uses a probabilistic ambiguity set defined as a ball of distributions that are close to the empirical distribution of the training set in the sense of the Wasserstein metric. We relax the DRO formulation into a regularized learning problem whose regularizer is a norm of the coefficient matrix. We establish out-of-sample performance guarantees for the solutions to our model, offering insights on the role of the regularizer in controlling the prediction error. Experimental results show that our approach improves the predictive error by 7\% -- 37\% for MLR, and a metric of robustness by 100\% for MLG.
\end{abstract}

\section{Introduction} \label{sec:intro}
We consider the multivariate learning problem under the framework of {\em
	Distributionally Robust Optimization (DRO)} where the ambiguity set is defined via
the Wasserstein metric \cite{gao2016distributionally, esfahani2018data}. The term
multivariate learning refers to scenarios where multiple correlated responses are to
be predicted - {\em Multivariate Linear Regression (MLR)}, or one of multiple classes
is to be assigned - {\em MultiClass Classification (MCC)}, based on a linear
combination of a set of predictors. Both involve learning a target vector $\by$ from
a vector of covariates $\bx$. We focus on developing \textbf{robust} multivariate
learning algorithms that are immunized against the presence of outliers in the data,
motivated by the fact that standard approaches, such as {\em Ordinary Least Squares
	(OLS)} and logistic regression (LG), are vulnerable to contamination of the dataset
by outliers. Robust models are desired in the scenarios where $(i)$ the training data
population differs significantly from the population to which the model shall be
applied, e.g., {\em covariate shift}
\cite{bickel2009discriminative}; and $(ii)$ we seek a model that works well over the entire data
range of interest; or (iii) we value model performance in the less frequently
occurring regions.

DRO, which minimizes the worst-case loss over a probabilistic ambiguity set, has received an increasing attention for inducing robustness to learning algorithms, due to its probabilistic interpretation of the uncertain data, tractability when assembled with certain metrics, and extraordinary performance observed on numerical examples. The ambiguity set in DRO can be defined through moment constraints \cite{goh2010distributionally, zymler2013distributionally, Sim14}, or as a ball of distributions using some probabilistic distance function such as the $\phi$-divergences
\cite{bayraksan2015data, hu2016does} and the Wasserstein distance, or a combination of both \cite{gao2017distributionally}.
The Wasserstein DRO model has been extensively studied in the machine learning community; see, for example, \cite{chen2018robust, blanchet2016robust, blanchet2019multivariate} for robustified regression models, \cite{sinha2017certifiable} for adversarial training in neural networks, and \cite{abadeh2015distributionally} for distributionally robust logistic regression. \cite{shafieezadeh2017regularization, gao2017wasserstein} provided a comprehensive analysis of the Wasserstein-based distributionally robust statistical learning problems with univariate response.

Most of the work on distributionally robust learning has focused on the univariate response scenario where the target $y$ is a scalar. In this paper, we extend this framework to the multiple response setting by exploring MLR with a Lipschitz continuous loss function and deriving the corresponding robust formulation. For the MCC problem, we incorporate the log-loss (negative log-likelihood) into the DRO formulation and derive the robust counterpart of multiclass logistic regression.

We adopt a DRO formulation that minimizes the worst-case expected loss within an ambiguity set that includes all probability distributions that are close to the nominal distribution in the sense of the Wasserstein metric. 
Unlike the univariate learning problem where the response variable is scalar and a coefficient vector representing the dependency of the response on the predictors is to be learned, in the multivariate setting the decision variable is a coefficient matrix $\bB \in \mbb{R}^{p \times K}$ whose $k$-th column explains the variation in the $k$-th coordinate of $\by \in \mbb{R}^K$ that can be attributed to the predictors $\bx \in \mbb{R}^p$, for $k=1,\ldots,K$. Inspired by the DRO relaxation derived in \cite{chen2018robust} for the univariate case, which adds a dual norm regularizer to the empirical loss, we obtain a novel \textbf{matrix norm} regularizer for the multivariate case through reformulating the Wasserstein DRO problem; thus, establishing a connection between robustness and regularization. This matrix norm exploits the geometrical structure of the coefficient matrix, and provides a way of associating the coefficients for the potentially correlated responses through the dual norm of the distance metric in the data space, enabling a primal-dual interpretation for the data-coefficient relationship. Note that the link between robustness and regularization has been established in the univariate learning setting, see, e.g., \cite{LAU97,el2003robust,xu2009robustness, xu2009robust, bertsimas2017characterization} for deterministic disturbances, \cite{lam2016robust, duchi2016statistics, gotoh2018robust} for stochastic disturbances within a $\phi$-divergence based ambiguity set, and \cite{abadeh2015distributionally, chen2018robust, blanchet2016robust, blanchet2017distributionally, blanchet2019multivariate, shafieezadeh2017regularization, gao2017wasserstein} for disturbances within a Wasserstein set. However, none of these works studied the multivariate robust problem.

To the best of our knowledge, we are the first to study the robust multivariate learning problem from the standpoint of Wasserstein distributional robustness. The extension from univariate DRO to the multivariate case is non-trivial. Our formulations are derived by analyzing the fundamental min-max problem. They are not simple superpositions of $K$ univariate relaxations and in general cannot be decomposed. Our method essentially provides a systematic and efficient way of integrating the individual coordinates of $\by$, instead of simply superposing them. Our model is general enough to encompass a class of regularizers that are related to the distance metric in the data space; thus, establishing a connection between robustness and regularization in the multivariate scenario.
Our approach is completely optimization-based, without the need to explicitly model
the complicated relationship between different responses, leading to compact and
computationally solvable models. It is interesting that a purely optimization-based
method that is completely agnostic to the covariate and response correlation
structure can be used as a better-performing alternative to statistical approaches
that
explicitly model this correlation structure. 

The rest of the paper is organized as follows. In Section \ref{s2}, we
develop the Wasserstein DRO formulations for MLR and Multiclass Logistic Regression (MLG), and introduce the matrix norm regularizer. Section~\ref{s3}
establishes the out-of-sample performance guarantees for the DRO solutions. The numerical experimental results are presented in
Section \ref{s4}. We conclude the paper in Section \ref{s5}.

\paragraph{Notational convention.} We use boldfaced lowercase letters to
denote vectors, ordinary lowercase letters to denote scalars, boldfaced
uppercase letters to denote matrices, and calligraphic capital letters
to denote sets. All vectors are column vectors. For space
saving reasons, we write $\bx=(x_1, \ldots, x_{\text{dim}(\bx)})$ to
denote the column vector $\bx$, where $\text{dim}(\bx)$ is the dimension
of $\bx$. We use prime to denote the transpose, $\|\cdot\|_p$
for the $\ell_p$ norm with $p \ge 1$, and $\|\cdot\|$ for the general vector norm that satisfies the following properties: $(i)$ $\|\bx\| = 0$ implies $\bx=\mathbf{0}$; $(ii)$ $\|a\bx\| = |a| \|\bx\|$, for any scalar $a$; $(iii)$ $\|\bx + \by\| \le \|\bx\| + \|\by\|$; $(iv)$ $\|\bx\| = \||\bx|\|$, where $|\bx| = (|x_1|, \ldots, |x_{\text{dim}(\bx)}|)$; and $(v)$ $\|(\bx, \mathbf{0})\| = \|\bx\|$, for an arbitrarily long vector $\mathbf{0}$. Note that any {\small $\bW$}-weighted $\ell_p$ norm defined as {\small $\|\bx\|_p^{\bW} \triangleq \big((|\bx|^{p/2})'\bW |\bx|^{p/2}\big)^{1/p} $} with a positive definite matrix {\small $\bW$} satisfies the above conditions, where $|\bx|^{p/2} = (|x_1|^{p/2}, \ldots, |x_{\text{dim}(\bx)}|^{p/2})$. Finally, $\|\cdot\|_*$ denotes the dual norm of $\|\cdot\|$ defined as $\|\btheta\|_* \triangleq \sup_{\|\bz\|\le 1}\btheta'\bz$, and $\bI_{K}$ denotes the $K$-dimensional identity matrix. 

\section{Formulations} \label{s2}
In this section we introduce the Wasserstein DRO formulations for MLR and MLG, and offer a dual norm interpretation for the regularization terms using a newly defined matrix norm.

\subsection{Multivariate Linear Regression} \label{sec:lr}
We assume the following linear model for the MLR problem: 
$$\by = \bB'\bx + \boldsymbol{\eta},$$
where $\by = (y_1, \ldots, y_K)$ is the vector of $K$ responses, potentially correlated with each other; $\bx = (x_1, \ldots, x_p)$ is the vector of $p$ predictors; $\bB = (B_{ij})_{i=1,\ldots,p} ^{j = 1, \ldots, K}$ is the $p \times K$ matrix of coefficients, the $j$-th column of which describes the dependency of $y_j$ on the predictors; $\boldsymbol{\eta}$ is the random error. Note that this assumption does not restrict us to linear models. A nonlinear extension, e.g., kernel regression, can be considered as a transformation on the input variables $\bx$, and applying a linear model to the transformed variables is equivalent to constructing a nonlinear model on the original inputs. From this perspective a linear setting seems to be adequate. Suppose we observe $N$ realizations of the data, denoted by $(\bx_i, \by_i), i=1,\ldots,N$, where $\bx_i = (x_{i1}, \ldots, x_{ip}), \by_i = (y_{i1}, \ldots, y_{iK})$.  
The Wasserstein DRO formulation for MLR minimizes the following worst-case expected loss: 
\begin{equation} \label{dro} 
\inf\limits_{\bB}\sup\limits_{\mbb{Q} \in \Omega}
\mbb{E}^{\mbb{Q}} [h_{\bB}(\bx, \by)],
\end{equation}
where $h_{\bB}(\bx, \by) \triangleq l(\by-\bB'\bx)$, with $l: \mbb{R}^K \rightarrow \mbb{R}$ an $L$-Lipschitz continuous function on the metric spaces $(\scrD, \|\cdot\|_r)$ and $(\scrC, |\cdot|)$, where $\scrD, \scrC$ are the domain and codomain of $l(\cdot)$, respectively; and $\mbb{Q}$ is the probability distribution of the data $(\bx, \by)$, belonging to a set $\Omega$ defined as
\begin{equation*}
\Omega \triangleq \{\mbb{Q}\in \scrP(\scrZ):\ W_1(\mathbb{Q},\ \hat{\mathbb{P}}_N) \le \epsilon\},
\end{equation*}
where $\scrZ$ is the set of possible values for $(\bx, \by)$; $\scrP(\scrZ)$ is
the space of all probability distributions supported on $\scrZ$; $\epsilon$ is a pre-specified positive constant; $\hat{\mbb{P}}_N$ is the empirical distribution that assigns equal probability to each observed sample;
$W_1(\mbb{Q},\ \hat{\mbb{P}}_N)$ is the order-1 Wasserstein distance
between $\mbb{Q}$ and $\hat{\mbb{P}}_N$ defined as 
\begin{equation} \label{wass1} 
W_1 (\mbb{Q}, \ \hat{\mbb{P}}_N) \triangleq \min\limits_{\Pi \in \scrP(\scrZ \times \scrZ)} \Bigl\{\int_{\scrZ \times \scrZ} s(\bz_1, \bz_2) \ \Pi \bigl(d\bz_1, d\bz_2\bigr)\Bigr\}, 
\end{equation}
where $\bz_i = (\bx_i, \by_i), i=1,2$, $\Pi$ is the joint distribution of $\bz_1$ and $\bz_2$ with
marginals $\mbb{Q}$ and $\hat{\mbb{P}}_N$, respectively, and $s(\cdot, \cdot)$ is a distance metric on the data space that measures the cost of transporting the probability mass. In the regression setting we define $s(\bz_1, \bz_2) \triangleq \| \bz_1 - \bz_2\|_r$. Notice that we use the same norm to define the Wasserstein metric and the metric space on the domain $\scrD$ of $l(\cdot)$. Problem (\ref{dro}) is difficult to work with due to the intractable high-dimensional integrals in the objective function and thus, a tractable relaxation is needed. We present a reformulation of (\ref{dro}) in Theorem \ref{thm1}. The proof can be found in the Supplementary.

\begin{thm} \label{thm1}
	Suppose we observe $N$ realizations of the data, denoted by $(\bx_i, \by_i), i=1,\ldots,N$. When the Wasserstein metric is induced by $\|\cdot\|_r$,
	the DRO problem (\ref{dro}) can be relaxed to:
	\begin{equation} \label{relax1}
	\inf_{\bB} \frac{1}{N}\sum_{i=1}^N h_{\bB}(\bx_i, \by_i) + \epsilon L \Big(\sum_{i=1}^K \|\bb_i\|_s^r\Big)^{1/r},
	\end{equation}
	and,
	\begin{equation} \label{relax2}
	\inf_{\bB} \frac{1}{N}\sum_{i=1}^N h_{\bB}(\bx_i, \by_i) + \epsilon L \|\bv\|_s,
	\end{equation}
	where $r, s \ge 1$, $1/r+1/s=1$, $\bb_i = (-B_{1i}, \ldots, -B_{pi}, \mathbf{e}_i)$ is the $i$-th row of $\tilde{\bB}$, with $\mathbf{e}_i$ the $i$-th unit vector in $\mbb{R}^K$, and $\bv \triangleq (v_1, \ldots, v_p, 1, \ldots, 1)$, with $v_i = \sum_{j=1}^K |B_{ij}|$, i.e., $v_i$ is a condensed representation of the coefficients for predictor $i$ through summing over the $K$ coordinates. We call (\ref{relax1}) the MLR-SR relaxation, and (\ref{relax2}) the MLR-1S relaxation (the naming convention will be more clear after introducing the $L_{r,s}$ matrix norm in Section \ref{matrixnorm}).  
\end{thm}

The regularization term in (\ref{relax1}) penalizes the aggregate of the dual norm of the regression coefficients corresponding to each of the $K$ responses. Notice that when $r \neq 1$, (\ref{relax1}) cannot be decomposed into $K$ independent terms. When $s=r=2$, the regularizer is just the Frobenius norm of $\tilde{\bB}$. The MLR-1S relaxation (\ref{relax2}) cannot be decomposed into $K$ subproblems when $s \neq 1$, due to the entangling of coefficients in the regularization term. 

Note that when $K=1$, with a 1-Lipschitz continuous loss, the regularizers in MLR-SR and MLR-1S reduce to $\epsilon \|(-\bbeta, 1)\|_s$, which coincides with the Wasserstein DRO formulation derived in \cite{chen2018robust}.
In both relaxations for MLR, the Wasserstein ball radius $\epsilon$ and the Lipschitz constant $L$ determine the strength of the penalty term. Recall that we assume the loss function is Lipschitz continuous on the same norm space with the one used by the Wasserstein metric. This assumption can be relaxed by allowing a different norm space for the Lipschitz continuous loss function, and the derivation technique can be easily adapted to obtain relaxations to (\ref{dro}). On the other hand, however, the norm space used by the Wasserstein metric can provide implications on what loss function to choose. For example, if we restrict the class of loss functions $l(\cdot)$ to the norms, our assumption suggests that $l(\bz) = \|\bz\|_r$, which is a reasonable choice since it reflects the distance metric on the data space.

\subsection{A New Perspective on the Formulation} \label{matrixnorm}

We will present a matrix norm interpretation for the two relaxations (\ref{relax1}) and (\ref{relax2}). Different from the commonly used matrix norm definitions in the literature, e.g., the vector norm-induced matrix norm $\|\bA\| \triangleq \max_{\|\bx\| \le 1} \|\bA \bx\|$, the entrywise norm that treats the matrix as a vector, and the Schatten norm that defines the norm on the vector of singular values, we adopt the $L_{r,s}$ matrix norm, which summarizes each column by its $\ell_r$ norm, and then computes the $\ell_s$ norm of the aggregate vector. The formal definition is described as follows.

\begin{defi}[$L_{r,s}$ Matrix Norm]
	For any $m \times n$ matrix $\bA = (a_{ij})_{i=1,\ldots,m}^{j=1,\ldots,n}$, define its $L_{r,s}$ norm as:
	\begin{equation*}
	\|\bA\|_{r,s} \triangleq \Bigg(\sum_{j=1}^n\bigg(\sum_{i=1}^m |a_{ij}|^r\bigg)^{s/r}\Bigg)^{1/s},
	\end{equation*}
	where $r, s \ge 1$. 
\end{defi}
Note that $\|\bA\|_{r,s}$ can be viewed as the $\ell_s$ norm of a newly defined vector $\bv = (v_1, \ldots, v_n)$, where $v_j = \|\bA_j\|_r$, with $\bA_j$ the $j$-th column of $\bA$. When $r=s=2$, the $L_{r,s}$ norm is the Frobenius norm. Moreover, $\|\bA\|_{r,s}$ is a convex function in $\bA$. 
The $L_{r,s}$ matrix norm depends on the structure of the matrix, and transposing a matrix changes its norm. For example, given $\bA \in \mbb{R}^{n \times 1}$, $\|\bA\|_{r,s} = \|\ba\|_r$, $\|\bA'\|_{r,s} = \|\ba\|_s$, where $\ba$ represents the vectorization of $\bA$. 

We can show that the $L_{r,s}$ norm is a valid norm (see the Supplementary). Moreover, it satisfies the following {\em sub-multiplicative} property:
\begin{equation*}
\|\bA \bB\|_{r,s}  \le \|\bA\|_{1, u} \|\bB\|_{t, s},
\end{equation*}
for $\bA \in \mbb{R}^{m \times n}, \bB \in \mbb{R}^{n \times K}$, and any $t, u \ge 1$ satisfying $1/t+1/u=1$.

Next we will rewrite the two relaxations (\ref{relax1}) and (\ref{relax2}) using the $L_{r,s}$ norm. When the Wasserstein metric is defined by $\|\cdot\|_r$, the MLR-SR relaxation can be written as:
\begin{equation*} 
\inf_{\bB} \frac{1}{N}\sum_{i=1}^N h_{\bB}(\bx_i, \by_i) + \epsilon L \|\tilde{\bB}'\|_{s, r}.
\end{equation*}
Similarly, the MLR-1S relaxation can be written as:
\begin{equation*} 
\inf_{\bB} \frac{1}{N}\sum_{i=1}^N h_{\bB}(\bx_i, \by_i) + \epsilon L \|\tilde{\bB}\|_{1, s},
\end{equation*}
where $r, s \ge 1$ and $1/r+1/s=1$. When the loss function is convex, e.g., $h_{\bB}(\bx, \by) = \|\by-\bB'\bx\|$, it is obvious that both MLR-SR and MLR-1S are convex optimization problems. By using the $L_{r,s}$ matrix norm, we are able to express the two relaxations in a compact way, which reflects the role of the norm space induced by the Wasserstein metric on the regularizer, and demonstrates the impact of the size of the Wasserstein ambiguity set and the Lipschitz continuity of the loss function on the regularization strength.

\subsection{Multiclass Logistic Regression} \label{sec:lg}
In this subsection we will apply Wasserstein DRO to the MLG problem. Suppose there are $K$ classes, and we are given a predictor vector $\bx \in \mbb{R}^p$. Our goal is to predict its class label, denoted by a $K$-dimensional binary label vector $\by \in \{0, 1\}^K$, where $\by = (y_1, \ldots, y_K)$, $\sum_k y_k = 1$, and $y_k=1$ if and only if $\bx$ belongs to class $k$. The conditional distribution of $\by$ given $\bx$ is modeled as
$p(\by|\bx) = \prod_{i=1}^K p_i^{y_i},$
where $p_i = e^{\bw_i'\bx}/\sum_{k=1}^K e^{\bw_k'\bx}$, and $\bw_i, i=1,\ldots, K$, are the coefficient vectors to be estimated that account for the contribution of $\bx$ in predicting the class labels.  The log-likelihood can be expressed as:
\begin{equation*}
\log p(\by|\bx)  = \sum_{i=1}^K y_i \log(p_i) 
= \by'\bB'\bx - \log \mathbf{1}'e^{\bB'\bx},
\end{equation*} 
where $\bB \triangleq [\bw_1 \cdots \bw_K]$, $\mathbf{1}$ is the vector of ones, and the exponential operator is applied element-wise to the exponent vector. The log-loss is defined to be the negative log-likelihood, i.e., $h_{\bB}(\bx, \by) \triangleq \log \mathbf{1}'e^{\bB'\bx} - \by'\bB'\bx$. The Wasserstein DRO formulation for MLG minimizes the following worst-case expected loss: 
\begin{equation} \label{dro-mlg} 
\inf\limits_{\bB}\sup\limits_{\mbb{Q} \in \Omega}
\mbb{E}^{\mbb{Q}} \Big[\log \mathbf{1}'e^{\bB'\bx} - \by'\bB'\bx \Big],
\end{equation}
where $\mbb{Q}$ and $\Omega$ are defined in the same way as in Section \ref{sec:lr}. We use the following distance function to define the Wasserstein metric: 
\begin{equation} \label{lg-dist}
s(\bz_1, \bz_2) = \|\bx_1 - \bx_2\|_r + Ms_{\by}(\by_1, \by_2),
\end{equation}
where $\bz_1 = (\bx_1, \by_1), \ \bz_2 = (\bx_2, \by_2)$, $s_{\by}(\cdot, \cdot)$ could be any metric, and M is a very large positive constant. In Theorem \ref{thm2} we derive a tractable relaxation of (\ref{dro-mlg}) by analyzing the growth rate of the log-loss function. The proof can be found in the Supplementary.
\begin{thm} \label{thm2}
	Suppose we observe $N$ realizations of the data, denoted by $(\bx_i, \by_i), i=1,\ldots,N$. When the Wasserstein metric is induced by (\ref{lg-dist}),
	the DRO problem (\ref{dro-mlg}) can be relaxed to:
	{\small 	\begin{equation} \label{lg-relax1}
		\inf_{\bB} \frac{1}{N} \sum_{i=1}^N \Bigl(\log \mathbf{1}'e^{\bB'\bx_i} - \by_i'\bB'\bx_i \Bigr) + \epsilon \Big(K^{1/s}  \|\bB\|_{s, r}+ \|\bB\|_{s, 1}\Big), 
		\end{equation}}
	and, 
	{\small 	\begin{equation} \label{lg-relax2}
		\inf_{\bB} \frac{1}{N} \sum_{i=1}^N \Bigl(\log \mathbf{1}'e^{\bB'\bx_i} - \by_i'\bB'\bx_i \Bigr) + \epsilon \Big(K^{1/s}  \|\bB'\|_{1,s}+ \|\bB\|_{s, 1}\Big), 
		\end{equation}}
	where $r, s \ge 1$, and $1/r+1/s=1$. We call (\ref{lg-relax1}) the MLG-SR relaxation, and (\ref{lg-relax2}) the MLG-1S relaxation. 
\end{thm}
Note that both MLG-SR and MLG-1S are convex optimization problems. When $K=2$, by taking one of the two classes as a reference, we can set one column of $\bB$ to zero, in which case all three regularizers $\|\bB\|_{s, r}$, $\|\bB\|_{s, 1}$ and $\|\bB'\|_{1,s}$ reduce to $\|\bbeta\|_s$, where $\bB \triangleq [\bbeta, \mathbf{0}]$, and our MLG-SR and MLG-1S relaxations coincide with the regularized logistic regression formulation derived in Remark 1 of \cite{abadeh2015distributionally} as their parameter $\kappa$ tends to infinity.

We also note that the number of classes $K$, and the Wasserstein set radius $\epsilon$, determine the regularization magnitude in the two MLG relaxations. There are two terms in the regularizer, one accounting for the predictor/feature uncertainty, and the other accounting for the label uncertainty. In the MLG-SR regularizer, we summarize each column of $\bB$ by its dual norm, and aggregate them by the $\ell_r$ and $\ell_1$ norms to reflect the predictor and label uncertainties, respectively. 

\section{Performance Guarantees for MLG} \label{s3}
In this section we will show out-of-sample performance guarantees for the DRO solutions, i.e., given a new test sample, what is the expected prediction bias/log-loss for using our estimator. The resulting bounds shed light on the role of the regularizer in inducing a low prediction error. 
Due to the space limitation, we will only show the results for MLG. The results for MLR can be found in the Supplementary. We first make several assumptions that are needed to establish the results.

\begin{ass} \label{b1} The $\ell_r$ norm of the predictor $\bx$ is bounded above almost surely, i.e., $\|\bx\|_r \le R_{\bx}.$
\end{ass}

\begin{ass} \label{b2} For any feasible solution $\bB$ to MLG-SR:
	$K^{1/s}  \|\bB\|_{s, r}+ \|\bB\|_{s, 1} \le \bar{C}_{s,r}$.
\end{ass}

\begin{ass} \label{b3} For any feasible solution $\bB$ to MLG-1S: 
	$K^{1/s}  \|\bB'\|_{1,s}+ \|\bB\|_{s, 1} \le \bar{C}_{1,s}$.
\end{ass}

With standardized predictors, $R_{\bx}$ in Assumption \ref{b1} can be assumed to be small. The form of the constraints in Assumptions \ref{b2} and \ref{b3} is consistent with the form of the regularizers in MLG-SR and MLG-1S, respectively. We will see later that the bounds $\bar{C}_{s,r}$ and $\bar{C}_{1,s}$ respectively control the out-of-sample log-loss of the solutions to MLG-SR and MLG-1S, which validates the role of the regularizer in improving the out-of-sample performance. 

\begin{thm} \label{c1} Suppose the solutions to MLG-SR and MLG-1S are $\hat{\bB}_{s,r}$ and $\hat{\bB}_{1,s}$, respectively. Under Assumptions \ref{b1} and \ref{b2}, for any
	$0<\delta<1$, with probability at least $1-\delta$ with respect to the
	sampling,
	\begin{equation*} 
		\begin{aligned}
		\mathbb{E}[\log \mathbf{1}'e^{\hat{\bB}_{s,r}'\bx} - \by'\hat{\bB}_{s,r}'\bx] & \le 
		\frac{1}{N}\sum_{i=1}^N
		(\log \mathbf{1}'e^{\hat{\bB}_{s,r}'\bx_i} - \by_i'\hat{\bB}_{s,r}'\bx_i) 
		+\frac{2 (R_{\bx} \bar{C}_{s,r} + \log K)}{\sqrt{N}} \\
		 & \quad +  
		(R_{\bx} \bar{C}_{s,r} + \log K)\sqrt{\frac{8\log(\frac{2}{\delta})}{N}}\ .
		\end{aligned}
		\end{equation*}
	Under Assumptions \ref{b1} and \ref{b3}, for any
	$0<\delta<1$, with probability at least $1-\delta$ w.r.t. the
	sampling,
	\begin{equation*} 
		\begin{aligned}
		\mathbb{E}[\log \mathbf{1}'e^{\hat{\bB}_{1,s}'\bx} - \by'\hat{\bB}_{1,s}'\bx] & \le
		\frac{1}{N}\sum_{i=1}^N
		(\log \mathbf{1}'e^{\hat{\bB}_{1,s}'\bx_i} - \by_i'\hat{\bB}_{1,s}'\bx_i)  
		+\frac{2 (R_{\bx} \bar{C}_{1,s} + \log K)}{\sqrt{N}} \\
		& \quad + 
		(R_{\bx} \bar{C}_{1,s} + \log K)\sqrt{\frac{8\log(\frac{2}{\delta})}{N}}\ .
		\end{aligned}
		\end{equation*}
\end{thm}

\textbf{Remark:} The bounds derived in Theorem \ref{c1} depend on the parameters $R_{\bx}$, and $\bar{C}_{s,r}$ ($\bar{C}_{1,s})$, which implicitly depend on the problem dimension $p+K$. The expected log-loss on a new test sample depends both on the sample average log-loss on the training set, and the magnitude of the regularizer in the formulation. The form of the bounds in Theorem \ref{c1} demonstrates the validity of MLG-SR and MLG-1S in leading to a good out-of-sample performance. For $r \ge 2$, $\bar{C}_{s,r}$ can be considered smaller than $\bar{C}_{1,s}$, while for $r=1$, the reverse holds. We can decide which model to use on a case-by-case basis, by computing their out-of-sample error on a validation set.

\section{Numerical Results} \label{s4}
In this section, we will test the out-of-sample performance of the MLR and MLG
relaxations on a number of synthetic datasets, and compare against several commonly
used multivariate regression/classification models. We use gradient descent to solve
the corresponding optimization problems.

\subsection{MLR Relaxations} \label{mlr-exp}
We generate the data as follows. The predictor $\bx \sim \scrN(\mathbf{0}, \bSigma_{\bx})$, where $\bSigma_{\bx} = (\sigma^{\bx}_{ij})_{i,j=1,\ldots,p}$, with $\sigma^{\bx}_{ij} = 0.9^{|i-j|}$. The response $\by$ is generated as $\by = (\bB^*)'\bx + \boldsymbol{\eta}$, where $\bB^*$ is generated from a standard normal distribution, and $\boldsymbol{\eta}$ is a standard normal random vector. Throughout the experiments we set $p=5$ and $K=3$.
We adopt a loss function $h_{\bB}(\bx, \by) = \|\by-\bB'\bx\|_2$ that is 1-Lipschitz continuous on $\|\cdot\|_2$. The Wasserstein metric is also induced by $\|\cdot\|_2$. 

We compare our MLR-SR and MLR-1S formulations with several other popular methods for MLR, including \textsl{OLS}, \textsl{Reduced Rank Regression (RRR)} \cite{izenman1975reduced, velu2013multivariate}, \textsl{Principal Components Regression (PCR)} \cite{massy1965principal}, \textsl{Factor Estimation and Selection (FES)} \cite{yuan2007dimension}, the \textsl{Curds and Whey (C\&W)} procedure \cite{breiman1997predicting}, and \textsl{Ridge Regression (RR)} \cite{brown1980adaptive, haitovsky1987multivariate}. Please refer to the Supplementary for a brief overview of these methods. The performance metrics we use include the out-of-sample {\em Weighted Mean Square Error (WMSE)}:
$\text{WMSE} \triangleq (1/M) \sum_{i=1}^M (\by_i-\hat{\by}_i)' \hat{\bSigma}^{-1} (\by_i-\hat{\by}_i),$
where $M$ is the size of the test set, $\by_i$ and $\hat{\by}_i$ are the true and predicted response vectors for the $i$-th test sample, respectively, and $\hat{\bSigma}$ is the covariance matrix of the prediction error on the training set, i.e., $\hat{\bSigma} = (\bY-\hat{\bY})'(\bY-\hat{\bY})/(N-pK)$, where $\bY, \hat{\bY} \in \mbb{R}^{N \times K}$ are the true and estimated response matrices of the training set, respectively, and $N$ is the size of the training set. We will also consider the {\em Conditional Value at Risk (CVaR)} of the WMSE (at the confidence level $\alpha = 0.8$) that quantifies its tail behavior. 

To test the robustness and examine the generalization performance of various methods,
we test the models on data whose distribution differs from the training population. Specifically, we inject two types of outliers to the test datasets: (i) outliers in the response direction, where the input distribution stays unchanged, but the response of outliers is generated as $\by = (\bB^*)'\bx + \boldsymbol{\eta}+\mathbf{o}_{\boldsymbol{\eta}}$, where  $\mathbf{o}_{\boldsymbol{\eta}} \sim \mathcal{N}(\mathbf{0}, \bSigma_{\by})$, with $\bSigma_{\by} = (\sigma^{\by}_{ij})_{i,j=1,\ldots,K}$, and $\sigma^{\by}_{ij} = (-0.9)^{|i-j|}$; and (ii) outliers in the predictors (covariate shift), where the predictors of outliers are generated as $\bx \sim \scrN(\mathbf{0}, \bSigma_{\bx}) + \scrN(\mathbf{0}, \bSigma_{\bx}^{\text{noise}})$, where $\bSigma_{\bx}^{\text{noise}} = (\sigma^{\text{noise}}_{ij})_{i,j=1,\ldots,p}$, with $\sigma^{\text{noise}}_{ij} = (-0.5)^{|i-j|}$.

\begin{figure}[h]
	\begin{center}
		\begin{subfigure}[b]{0.49\columnwidth}
			\centering
			\includegraphics[width=\linewidth]{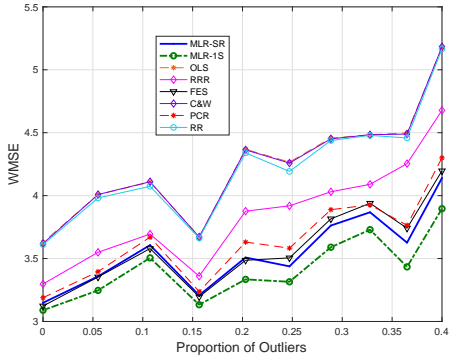}
			\caption{\small{WMSE.}}
		\end{subfigure}
		\begin{subfigure}[b]{0.49\columnwidth}
			\centering
			\includegraphics[width=\linewidth]{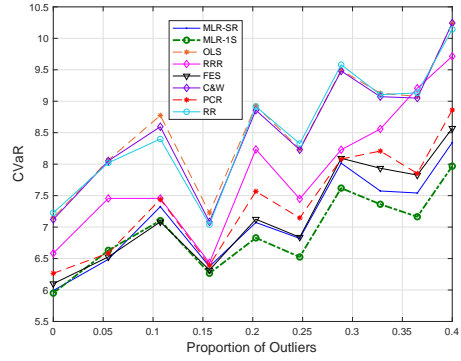}
			\caption{\small{CVaR of WMSE.}}
		\end{subfigure}
	\end{center}
	\vspace{-14pt}
	\caption{The out-of-sample performance of different MLR models when outliers are in the response.}
	\label{fig:mlr-1}
\end{figure}

\begin{figure}[h]
	\begin{center}
		\begin{subfigure}[b]{0.49\columnwidth}
			\centering
			\includegraphics[width=\linewidth]{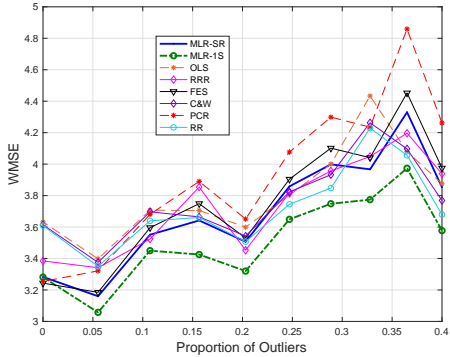}
			\caption{\small{WMSE.}}
		\end{subfigure}
		\begin{subfigure}[b]{0.49\columnwidth}
			\centering
			\includegraphics[width=\linewidth]{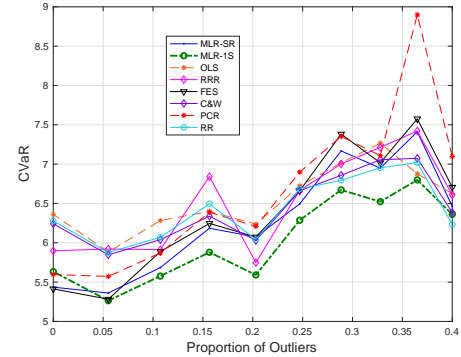}
			\caption{\small{CVaR of WMSE.}}
		\end{subfigure}
	\end{center}
	\vspace{-14pt}
	\caption{The out-of-sample performance of different MLR models when outliers are in the predictors.}
	\label{fig:mlr-2}
\end{figure}

We generate 10 datasets with a training size of 100 and a test size of 60, and compare the WMSE and CVaR of various models on the test set. All the regularization coefficients are tuned through cross-validation. 
Figures \ref{fig:mlr-1} and \ref{fig:mlr-2} show the WMSE and CVaR of WMSE for the two scenarios, as the proportion of outliers in the test dataset changes. It is clear that our MLR-1S model (green line) achieves the smallest prediction error. As the proportion of outliers increases, all models perform worse, but the advantage of the MLR-1S model becomes more prominent. We also study the impact of the training size on the generalization performance. Please refer to the Supplementary for details. 

\subsection{MLG Relaxations}
We study the performance of the two MLG relaxations, and compare them with a number of MLG variants on simulated datasets.
The predictor is drawn according to $\bx \sim \scrN(\mathbf{0}, \mathbf{I}_p)$. The label vector $\by \in \{0, 1\}^K$ is generated from a 
multinomial distribution with probabilities specified by the softmax normalization of $(\bB^*)'\bx+\boldsymbol{\eta}$, where $\boldsymbol{\eta} \sim \scrN(\mathbf{0},\mathbf{I}_K)$, and $\bB^*$ is generated from a standard normal distribution.
We set $p=5, K=3$, and conduct 10 simulation runs, each with a training size of 100 and a test size of 60. The performance metrics we use include: $(i)$ the average log-loss, $(ii)$ the {\em Correct Classification Rate (CCR)}, and $(iii)$ the {\em Conditional Value at Risk (CVaR)} (at the confidence level 0.8) of log-loss which computes the expectation of extreme log-loss values. The average performance metrics on the test set are reported. 

We will compare against $(i)$ Vanilla MLG which minimizes the empirical log-loss with no penalty term, $(ii)$ Ridge MLG which penalizes the trace of $\bB'\bB$ as in ridge regression, $(iii)$ LASSO MLG which penalizes the sum of absolute values of all entries in $\bB$, and $(iv)$ PCC MLG which converts the predictors into a set of linearly uncorrelated variables and applies logistic regression on the transformed variables.
In addition to the three performance metrics used earlier, we introduce another robustness measure that calculates the minimal perturbation needed to ``fool'' the classifier. For a given $\bx$ with label $k$, for any $j \neq k$, consider the following optimization problem:
\begin{equation} \label{perturb}
\begin{aligned}
\min_{\tilde{\bx}} & \quad \|\bx - \tilde{\bx}\|_1 \\
\text{s.t.} & \quad P_j(\tilde{\bx}) \ge P_k(\tilde{\bx}),\\
& \quad k = \arg\max_i P_i(\bx),
\end{aligned}
\end{equation}
where $P_i(\bx)$ denotes the probability of assigning class label $i$ to $\bx$, which is a function of the trained classifier. Problem (\ref{perturb}) measures the minimal perturbation distance (in terms of the $\ell_1$-norm) that is needed to change the label of $\bx$. Its optimal value evaluates the robustness of a given classifier in terms of the perturbation magnitude. The more robust the classifier, the larger the required perturbation to switch the label, and thus the larger the optimal value. We solve problem (\ref{perturb}) for every test point $\bx$ and any $j \neq k$, and take the minimum of the optimal values to be the {\em Minimal Perturbation Distance (MPD)} of the classifier.

We test the model performance on datasets with covariate shift, to mimic the real applications where the input data get perturbed (e.g., blurred images).  Specifically, the predictors of the outliers in the test datasets are generated as $\bx \sim \scrN(\mathbf{0}, \bI_p) + \scrN(\mathbf{0}, \bSigma_{\bx}^{\text{noise}})$, where $\bSigma_{\bx}^{\text{noise}} = (\sigma^{\text{noise}}_{ij})_{i,j=1,\ldots,p}$, with $\sigma^{\text{noise}}_{ij} = 0.7^{|i-j|}$. The conditional label distribution stays the same. 
Table \ref{tab:mlg-2} shows the average performance of various models over 10 runs. Our MLG-SR and MLG-1S models, both induced by $r=2$, achieve similar prediction performance to others, in terms of CCR and log-loss, but obtain a remarkably higher MPD value, improving over others by 100\%, indicating a significantly higher robustness to data disturbances. In particular, the MLG-1S improves over PCC MLG the CCR by 12\%, the log-loss by 13\%, and the CVaR by 16\%, empirically demonstrating the superiority of a purely optimization-based method to a method that explicitly models the correlation structure between predictors. Note that MLG-1S slightly loses to Ridge MLG in terms of CCR, but has a lower log-loss and a higher MPD.

\vspace{-1em}
\begin{table}[h]
\caption{The out-of-sample performance of different MLG models trained on datasets with 20\% outliers, mean (std.)} 
\label{tab:mlg-2}
\begin{center}
	\begin{tabular} {l| c c c c}
			\hline
			& CCR                   & Log-loss                 & CVaR                     & MPD          \\ \hline
			MLG-SR       & 0.65 (0.01)          & 0.81 (0.03)               & 1.29 (0.14)              & 0.02 (0.03)  \\ 
			MLG-1S       & 0.66 (0.02)          & 0.78 (0.02)               & 1.28 (0.16)              & 0.02 (0.01)\\ 
			Vanilla MLG  & 0.66 (0.03)          & 0.82 (0.06)              & 1.45 (0.25)              & 0.01 (0.01)  \\ 
			Ridge MLG    & 0.67 (0.02)          & 0.80 (0.04)               & 1.34 (0.22)              & 0.01 (0.003) \\  
			PCC MLG      & 0.59 (0.08)          & 0.90 (0.14)               & 1.52 (0.24)              & 0.01 (0.01) \\
			LASSO MLG    & 0.66 (0.02)           & 0.79 (0.03)               & 1.30 (0.21)              & 0.01 (0.01) \\
			\hline
		\end{tabular}
	\end{center}
\end{table}
		
\section{Conclusions} \label{s5}
We proposed a novel distributionally robust framework for {\em Multivariate Linear Regression (MLR)} and {\em Multiclass Logistic Regression (MLG)}, where the worst-case expected loss over a probabilistic ambiguity set defined by the Wasserstein metric is being minimized. By exploiting the special structure of the Wasserstein metric, we relax the min-max formulation to a regularized empirical loss minimization problem. The regularization term is a function of the $L_{r,s}$ norm of the coefficient matrix, establishing a connection between robustness and regularization in the multivariate setting. We provide both theoretical performance guarantees to our estimators, and empirical evidence showing that our models achieve a better predictive performance than others, and a significantly higher robustness to outliers.

\end{document}